\documentclass[unnumsec,webpdf,contemporary,large,namedate]{oup-authoring-template}

\usepackage{setspace}
\usepackage{makecell}
\usepackage{color,soul}

\graphicspath{{Fig/}}

\theoremstyle{thmstyleone}%
\theoremstyle{thmstyletwo}%
\theoremstyle{thmstylethree}%

\newcommand{\ie}{\textit{i}.\textit{e}.}

\begin{document}

\journaltitle{Cerebral Cortex}
\DOI{DOI HERE}
\copyrightyear{2022}
\pubyear{2019}
\access{Advance Access Publication Date: Day Month Year}
\appnotes{Paper}

\firstpage{1}


\title[Early Autism Diagnosis Scheme]{Early Autism Diagnosis based on Path Signature and Siamese Unsupervised Feature Compressor}

\author[1,2]{Zhuowen Yin\ORCID{0009-0003-9049-8323}}
\author[1,4]{Xinyao Ding}
\author[1,6,$\ast$]{Xin Zhang}
\author[3]{Zhengwang Wu}
\author[3]{Li Wang}
\author[5,6]{Xiangmin Xu}
\author[3,$\ast$]{Gang Li}

\authormark{Yin et al.}

\address[1]{\orgdiv{School of Electronics and Information Engineering}, \orgname{South China University of Technology}, \orgaddress{\postcode{510641}, \state{Guangzhou, Guangdong Province}, \country{China}}}
\address[2]{\orgdiv{Department of Bioengineering, School of Engineering and Applied Science}, \orgname{University of Pennsylvania}, \orgaddress{\postcode{PA 19104}, \state{Philadelphia}, \country{USA}}}
\address[3]{\orgdiv{Department  of  Radiology  and  Biomedical  Research  Imaging  Center}, \orgname{University  of  North  Carolina  at  Chapel  Hill}, \orgaddress{\postcode{NC 27599}, \state{Chapel  Hill}, \country{USA}}}
\address[4]{\orgname{The Affiliated Shenzhen School of Guangdong Experimental High School}, \orgaddress{\postcode{518100}, \state{Shenzhen, Guangdong Province}, \country{China}}}
\address[5]{\orgdiv{School of Future Technology}, \orgname{South China University of Technology}, \orgaddress{\postcode{510641}, \state{Guangzhou, Guangdong Province}, \country{China}}}
\address[6]{\orgname{Pazhou Lab}, \orgaddress{\postcode{510330}, \state{Guangzhou, Guangdong Province}, \country{China}}}

\corresp[$\ast$]{Corresponding author. \href{eexinzhang@scut.edu.cn}{eexinzhang@scut.edu.cn} (X. Zhang); \href{gang_li@med.unc.edu}{gang\_li@med.unc.edu} (G. Li)}

\received{Date}{0}{Year}
\revised{Date}{0}{Year}
\accepted{Date}{0}{Year}



\abstract{Autism Spectrum Disorder (ASD) has been emerging as a growing public health threat. Early diagnosis of ASD is crucial for timely, effective intervention and treatment. However, conventional diagnosis methods based on communications and behavioral patterns are unreliable for children younger than 2 years of age. Given evidences of neurodevelopmental abnormalities in ASD infants, we resort to a novel deep learning-based method to extract key features from the inherently scarce, class-imbalanced, and heterogeneous structural MR images for early autism diagnosis. Specifically, we propose a Siamese verification framework to extend the scarce data, and an unsupervised compressor to alleviate data imbalance by extracting key features. We also proposed weight constraints to cope with sample heterogeneity by giving different samples different voting weights during validation, and we used Path Signature to unravel meaningful developmental features from the two-time point data longitudinally. We further extracted machine learning focused brain regions for autism diagnosis. Extensive experiments have shown that our method performed well under practical scenarios, transcending existing machine learning methods and providing anatomical insights for autism early diagnosis.}
\keywords{Autism, Class imbalance, Deep learning, Magnetic resonance imaging, Path Signature}


\maketitle

\section{Introduction}

According to a report from the Centers for Disease Control and Prevention (CDC), 1 in 44 American children is affected by Autism Spectrum Disorder (ASD), which has increased 241\% from 2000 to 2018 \citep{maenner2021prevalence}, becoming a growing public health threat. Consensus has been made that early diagnosis would be critical for more effective ASD treatment, as ASD is an early-onset neurodevelopmental disorder \citep{fernell2013early}. However, aside from the rapid growth of young ASD subjects and its influence on children, early ASD diagnosis advancements remain stagnant. This is because, due to unclear physiological causes of ASD and the lack of biological diagnostic markers, ASD might not be diagnosed until the identification of the required social deficits and behavioral patterns at the age of 2-3 years  \citep{american2013diagnostic, baird2003diagnosis}, which may miss the optimal time for effective intervention \citep{fernell2013early}.

The first two postnatal years are a period of extremely dramatic brain development \citep{lyall2015dynamic, li2019computational, wang2019developmental, huang2022mapping}, which contains critical information for the early diagnosis of ASD \citep{hazlett2017early}. Since brain MR imaging is highly informative, it provides a noninvasive way for researchers to investigate brain anatomy and function during this period. In ASD studies of older children, brain MRI data reveal meaningful patterns and biomarkers to inform diagnosis \citep{chen2011structural, yamagata2019machine, plitt2015functional}. Compared to the deficiency of early behavioral evidence for ASD diagnosis, recent studies have also shown that early brain developmental abnormalities can be observed in structural brain magnetic resonance imaging (sMRI) of ASD infants before the first appearance of autistic behaviors \citep{hazlett2017early}. We are thus motivated to extract information from structural MRI data for infant autism diagnosis.

Despite the presence of group-level cortical developmental abnormalities in MRI data of autistic children, individual-level abnormalities are more complex and heterogeneous \\\citep{katuwal2015predictive}, posing significant challenges in imaging-based automatic diagnosis and prediction. Therefore, machine learning methods, especially deep learning methods, could be used to explore aberrant ASD patterns from brain MRI data and characterize ASD features in a data-driven manner \citep{chen2015diagnostic}. Traditional machine learning has been applied in individual-level classification and prediction in ASD studies. For example, \citet{zhang2018whole} and \citet{rane2017developing} used SVM to accomplish ASD classification with diffusion MRI tractography data and fMRI data respectively, while \citet{moradi2017predicting} predicted ASD severity based on cortical thickness with SVR.

Compared to traditional machine learning methods, deep learning has shown exceptional performance on medical image analysis tasks \citep{ronneberger2015u}, due to its strong power in representation learning. Using brain MRI data, various deep learning models are proposed for ASD diagnosis. \citet{sherkatghanad2020automated} designed a multi-channel Convolutional Neural Network (CNN) with different filter sizes to process the functional connectivity (FC) matrices computed from ABIDE rs-fMRI dataset and obtained favorable results on multi-site data. \citet{iidaka2015resting} implemented a Probabilistic Neural Network (PNN) for ASD classification, based on functional connectivity matrices as well. Since the connections of brain regions constitute a graph, Graph Convolutional Networks (GCNs) have natural advantages for learning effective representations. \citet{arya2020fusing} proposed a 3D CNN-GCN method, where the CNN modules are used to extract fMRI low-dimensional feature vectors, and GCN is implemented for the further classification of the constructed feature graph. These works have taken the forms of input data into consideration, designing feature extraction methods that fit to single-subject properties. Despite their success, those models failed to establish a comprehensive scheme to consider more practical issues in neuroimage-based ASD diagnosis, like the small size of datasets and the categorical imbalance of data under realistic scenarios during infancy, due to the huge difficulties in the acquisition of pediatric neuroimaging. 

On the one hand, the small size of available training data-sets can limit the generalization ability of the trained deep learning models \citep{ronneberger2015u}, especially for neuroimages, which typically exhibit large inter-subject variability. To overcome the influence of small datasets and extract useful features, \citet{heinsfeld2018identification} utilized an autoencoder structure to extract key features for ASD classification. \citet{hazlett2017early} also adapted stacked autoencoders for feature compression, which concatenated longitudinal sMRI data as input feature vector and obtained fair classification results using SVM on the extracted features. Instead of feature extraction improvement, \citet{eslami2019asd} implemented a data augmentation method based on linear interpolation to produce more synthetic data to compensate for small training datasets. Although these methods focused on data augmentation or reconstruction to extract key features from small datasets, they ignored the heterogeneity between data samples, and the complexity of brain MRI data impeded feature extraction from small datasets.

On the other hand, the categorical balance of data is a strong assumption and can limit the practical application of trained models. Most existing methods assume the training and testing datasets are categorically balanced, in which the numbers of ASD participants are approximately equal to those of the normal controls \citep{rane2017developing, sherkatghanad2020automated, iidaka2015resting, heinsfeld2018identification, rathore2019autism, zhang2022detection}. However, this does not reflect the fact that ASD samples are typically minorities among all test samples in realistic scenarios \citep{knopf2020autism}. \citet{aghdam2018combination} used Deep Belief Network (DBN) to process imbalanced ABIDE fMRI and sMRI data consisting of 116 individuals with ASD and 69 typical controls, obtaining an accuracy of 65.56\% and sensitivity of only 32.96\%, which has shown the difficulties of deep feature extraction from imbalanced data. Among works with imbalanced data, \citet{hazlett2017early} conducted experiments based on imbalanced data by proposing a median network with feature binarization to obtain a more robust result. \citet{li2018early} adapted the data-driven attention among brain regions and adopted a data reuse strategy for imbalanced data, which alleviated the imbalance between classes. However, these works still cannot well cope with practical issues like small data size, sample heterogeneity, and data complexity.

To address the limitations of existing methods, we propose a novel class-imbalanced deep learning scheme based on Siamese verification and unsupervised feature compression for ASD diagnosis from sMRI data during infancy. The main contributions of our work are summarized below: 
\begin{enumerate}[(a)]
\item To overcome the limitation of small samples of subjects, a Siamese verification model is proposed. The traditional single-sample classification input is thus converted into a two-sample paired input, which leads to meaningful data enlargement. 
\item To address the problem of imbalance between sample categories, an unsupervised feature compression model based on a dual-channel stacked autoencoder is proposed to extract key features without the influence of imbalanced sample distribution. 
\item To deal with the heterogeneity among samples, we obtain different weights for different reference samples in Siamese verification vote regarding their similarity to the to-be-tested sample during the test phase, thus similar reference samples are given more importance. 
\item To enrich the feature representation and extract cortical properties from the complex longitudinal data more effectively, we compute the path signature (PS) of sMRI data and concatenate the obtained feature with the original input. 
\item Based on our proposed deep learning scheme, we further analyzed the focused brain regions of our model for early autism diagnosis, which are in line with evidences from other studies.
\end{enumerate}

\begin{figure*}
	\centering
		\centerline{\includegraphics[width=1.05\textwidth]{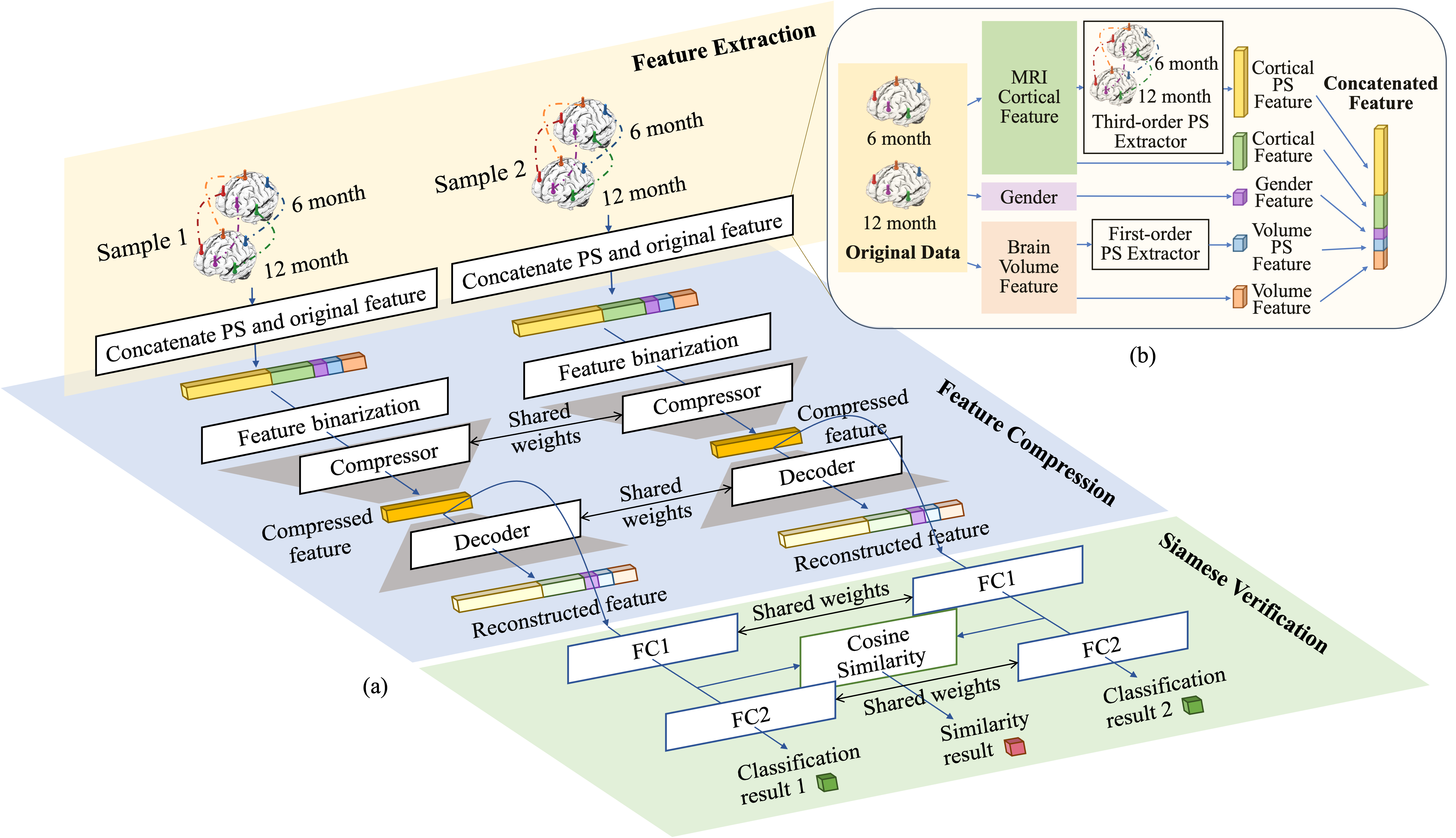}}
	\caption{
(a) The whole algorithm flow of the training process, which is divided into three parts: feature extraction, feature compression, and Siamese verification. In the feature extraction part, we extract longitudinal Path Signature of the features and concatenate them with the original morphological features. In the feature compression part, we train stacked dual auto-encoders (the Compressor) to compress and exploit useful features. In the Siamese verification part, we train a multi-task learning-based Siamese verification model, to ensure the validity of classification results. (b) The detailed structure of concatenating input features with PS features.}
	\label{whole-fig}
\end{figure*}

\section{Materials}
We verify the effectiveness of our proposed model using the T1w and T2w brain MR images gathered in the Autism Centers of Excellence Network study funded by the NIH, which is referred to as the Infant Brain Imaging Study (IBIS). Our dataset includes 30 autistic infants (23 males/7 females) and 127 normal infants (102 males/55 females). The recruitment and imaging are conducted at one of the four sites: University of North Carolina at Chapel Hill, University of Washington, Children’s Hospital of Philadelphia, and Washington University in St. Louis, as described in  \citet{wang2022developmental}. Study protocols are approved by the Institutional Review Boards of their sites respectively, and all participants' parents or legal guardians provided informed consent to participate. All images were acquired at around 6 and 12 months of age on Siemens 3T scanners. Subjects were naturally sleeping with their heads secured in a vacuum-fixation device and ear protected. T1w MR images were acquired with 160 sagittal slices using parameters: TR/TE=2400/3.16ms and voxel resolution = $1\times1\times1$ $mm^3$. T2w MR images were obtained with 160 sagittal slices using parameters: TR/TE = 3200/499 ms and voxel resolution = $1\times1\times1$ $mm^3$ \citep{hazlett2017early}. 

All infant MR images are processed using an established infant-specific computational pipeline iBEAT V2.0 \citep{wang2023ibeat} (www.ibeat.cloud) to reconstruct the cortical surfaces and generate the vertex-wise cortical features, including the surface area and the cortical thickness. Then, the reconstructed cortical surfaces are further parcellated into regions of interest (ROIs) by aligning onto the 4D infant cortical surface atlas \citep{li2015construction, wu2019construction} using the spherical demons \citep{yeo2009spherical}. Of note, the 4D atlas has several different parcellation strategies, with each strategy having different numbers of ROIs. We used 35 ROIs for each hemisphere, following the Freesurfer Desikan parcellation protocol. For each ROI, we extract two representative morphological features, the total surface area, and the average cortical thickness. In addition, for each subject at each time point, we also compute their total brain volume based on brain segmentation. Therefore, for each subject with two longitudinal time points (6 and 12 months), we totally have a 283-dimensional feature vector that includes 140 6-month ROI-wise cortical features, 140 12-month ROI-wise features, 2 total volume features, and the gender tag.
 
\section{Method}
Our method focused on realistic scenarios of ASD diagnosis in early stages, where the available infant MRI samples are usually of small numbers and categorically imbalanced. Regarding those features of data, we proposed a class-imba-lanced deep learning scheme based on multi-task learning and Siamese verification. The framework of our method is depicted in Fig. \ref{whole-fig}, which can be divided into three parts. The first part of our scheme is Path Signature based longitudinal feature extraction and pre-processing (“Feature Extraction” part in Fig. \ref{whole-fig}(a)), which explicitly extracts dynamics of two-timepoint data, extending feature dimensions to make these features easier to learn. The second part is the dual-channel autoencoder for feature compression (“Feature Compression” part in Fig. \ref{whole-fig}(a)), focusing on the extraction of key features to alleviate noises and imbalance. The third part is the multi-task learning based Siamese verification (“Siamese Verification” part in Fig. \ref{whole-fig}(a)), serving as data augmentation during training for improving the classification performance with small sample size. Each part will be detailed below.

\subsection{Path signature based longitudinal feature extraction and pre-processing}
The structural MRI data that we extracted and utilized from the NDAR dataset includes three types of features: the surface area and cortical thickness of 70 brain regions, the total brain volume, and the gender, collectively representing the morphological features of the infant brain cortex. Besides these features, we also want to capture the informative longitudinal developmental patterns of the cortex, by using both 6-month and 12-month sMRI data of the same subject. Specifically, we leverage Path Signature (PS) method to capture the longitudinal developmental patterns, which corresponds to the “Feature Extraction” part in Fig. \ref{whole-fig}(a). 

\subsubsection{Preliminary of Path Signature}
\label{sec:ps}
We briefly introduce the mathematical definition and geometric interpretation of path signature (PS), which mainly refers to \citet{chevyrev2016} as a mathematical feature extraction and expansion tool. Assume a path $P:[t_1, t_2]\to\mathbb{R}^d$, where $[t_1, t_2]$ is a time interval. The features of the path at the time point $t$ are denoted with $d$ dimensions as $(P_t^1,...,P_t^d)$, where on the dimension $i$, $P^i:[t_1, t_2]\to\mathbb{R}$ is a real-value path in the whole time interval. For an integer $k\ge1$ and the collection of indices $i_1,...,i_k\in\{1,...,d\}$, the \emph{k}-fold iterated integral of the path along indices $i_1,...,i_k$ can be defined as:
\begin{equation}
\begin{split}
S(P)_{t_1,t_2}^{i_1,...,i_k}=\int_{t_1<a_k<t_2}...\int_{t_1<a_1<a_2}dP_{a_1}^{i_1}...dP_{a_k}^{i_k}
\end{split}
\label{eq1}
\end{equation}
where $t_1<a_1<a_2<...<a_k<t_2$.

The signature of path $P$, denoted by $S(P)_{t_1,t_2}$, is the collection (infinite series) of all the iterated integrals of $P$:
\begin{equation}
\begin{split}
S(P)_{t_1,t_2}=&(1,S(P)_{t_1,t_2}^1,S(P)_{t_1,t_2}^2,...,S(P)_{t_1,t_2}^d,\\
&S(P)_{t_1,t_2}^{1,1},...,S(P)_{t_1,t_2}^{1,d},...,S(P)_{t_1,t_2}^{d,d},\\
&...,\\
&S(P)_{t_1,t_2}^{1,...,1},...,S(P)_{t_1,t_2}^{i_1,...,i_k},...,S(P)_{t_1,t_2}^{d,...,d},\\
&...)
\end{split}
\label{eq2}
\end{equation}

The \emph{k}-th level PS is the collection (finite series) of all the \emph{k-fold iterated integral} of path $P$. Intuitively, the \emph{1}-st and \emph{2}-nd level represent the path displacement and the signed area enclosed by the path respectively. By increasing level $k$, higher levels of path information can be extracted, but the dimensionality of iterated integrals grows rapidly as well. In practice, we usually truncate the $S(P)_{t_1,t_2}$ at a specific level $m$ to constrain the dimensionality of the PS feature in a reasonable range. 

\subsubsection{Concatenated longitudinal and morphological features}

We first compute the PS features along the longitudinal (temporal) direction, in order to capture the developmental patterns of the brain. Specifically, we define a path $P:[t_1,t_2]$ that starts at 6-month $t_1$ and ends at 12-month $t_2$. Hence, the PS cortical features (thickness and area) can be computed according to Equation \ref{eq2}, where $(P_{t_1}^1,...,P_{t_1}^d)$ and $(P_{t_2}^1,...,P_{t_2}^d)$ are morphological features at timepoint $t_1$ and $t_2$ respectively. Since the morphological features are high-dimensional, we set the truncate level $k=3$, so that PS features can be prevented from becoming too long. Similarly, we compute the PS feature of the total brain volume, where we set the truncate level $k=1$, which would be adequate to extract the developmental information of the single-dimensional total volume.

We concatenate all those longitudinal and morphological features mentioned above with the gender information to produce the concatenated input feature $F^{cat}$, as input for our dual channel autoencoder. Specifically, $F^{cat}$ is composed of five parts, \ie, the original cortical features, cortical PS features, total volume, volume PS feature, and gender, as shown in Fig. \ref{whole-fig}(b).

\subsection{Dual-channel autoencoder for feature compression}

After longitudinal feature extraction and feature concatenation, the raw input feature $F^{raw}$ is dimensionally expanded into concatenated feature $F^{cat}$, which is in high dimension but less representative. We therefore use a dual-channel autoencoder to exploit characteristic information of brain morphology and development. With binarized features and trained stacked autoencoders, key features can be compressed and extracted from imbalanced data to alleviate the influence of noise and imbalance. This section corresponds to the “Feature Compression” part of Fig. \ref{whole-fig}(a).

\subsubsection{Feature binarization}

In spite of information loss, input binarization is a direct way to denoise and avoid overfitting. The idea of input feature binarization was first introduced by \citet{rastegari2016xnor}, where the authors have showed that input binarization is an effective preprocessing step that only brings minor performance decrease. Feature binarization is then further used in small sample size learning scenarios, like in \citet{hazlett2017early}. The rationale of doing so is to avoid overfitting on small datasets by reducing data complexity, as discussed in \citet{bejani2021systematic}. Therefore, we perform binarization on path-signature-extracted features and then use them as input data of the dual-channel autoencoder network for further learnable feature compression.  Specifically, given subjects $i$ and $j$ and their concatenated features, $n_i$ and $n_j$, we have the input pair $F_{i,j}^{cat}=\{n_i,n_j\}$. For each feature dimension, we extract the median as the binarization threshold. Feature values larger than the median are set to 1, and others are set to 0. After binarization, the data pair is denoted as $F_{i,j}^{bin}=\{b_i,b_j\}$. During this step, the noises are smoothed out. We will show in the experiments that we achieve better performance with binarization. 

\subsubsection{Dual channel unsupervised feature compressor}

To alleviate the influence of imbalanced data, we utilize an unsupervised approach to obtain compressed feature vectors without imbalanced labels. Obtaining compressed compact developmental and structural features produced by stacked encoders also further removes data noises. Moreover, we adopt the dual channel structure for autoencoders, to make the training process consistent with the next network part. In general, we design a dual channel autoencoder-based unsupervised network to further extract the low-dimensional core features after binarization, which is called the Compressor, as shown in Fig. \ref{whole-fig}(a).

 Specifically, given binarized feature pair $F_{i,j}^{bin}=\{b_i,b_j\}$, we use them as the input of the dual channel compressor. As shown in Fig. \ref{fig-siamese}, for each channel, we design a two-layer encoder and a two-layer decoder, with their weights shared between channels. The compressed low-dimensional feature is obtained as the output of the encoder, which is referred to as the Compressor in Fig. \ref{whole-fig}(a). The loss function is the MSE loss between the original and reconstructed data, defined as, 

\begin{equation}
    \mathcal{L}_{recon}=\text{MSE}(F_{i,j}^{bin},F_{i,j}^{rec})=\frac{1}{n}(\sum_{i=1}^n(b_i-r_i)^2+\sum_{i=1}^n(b_j-r_j)^2)
    \label{recon_loss}
\end{equation}
where $n$ is the batch size, $F_{i,j}^{bin}=\{b_i,b_j\}$ is the original paired input feature, and $F_{i,j}^{rec}=\{r_i,r_j\}$ is the corresponding reconstructed paired feature. 

\begin{figure}[htb]
	\centering
	\centerline{\includegraphics[width=8.6cm]{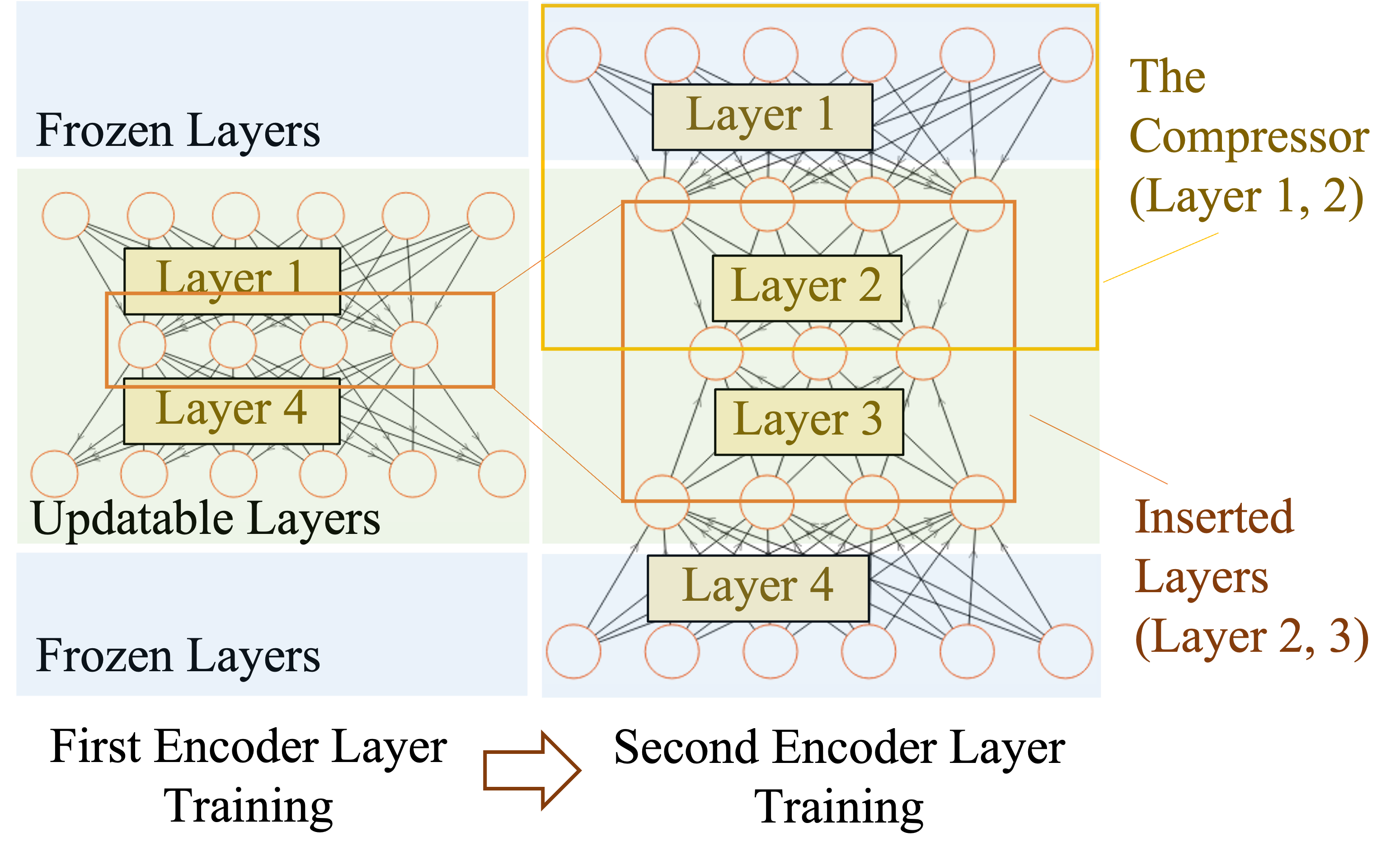}}
	\caption{ 
 Schematic diagram of the autoencoder training process introduced in Algorithm 2. We train layers 1 and 4 first, then freeze them, and insert layers 2 and 3 to train. We ultimately obtain layers 1 and 2 as the feature Compressor for further training, which outputs compressed low-dimensional feature vectors.  }
	\label{fig:autoencoder}
\end{figure}

To make the auto-encoder converge with small sample sizes, we propose a hierarchical training process for the feature compressor, as shown in Fig. \ref{fig:autoencoder} and Algorithm \ref{alg:Framework}. The hierarchical training process decomposes the whole-encoder optimization into two stages, which narrows the optimization search space. Specifically, our trained auto-encoder has four layers in total. At the starting point, only two layers (layer 1 and layer 4) are utilized and trained. After training those two layers in the first training stage, we freeze the parameters of layer 1 and layer 4, and then insert two new layers (layer 2 and layer 3) between layer 1 and 4. We then train layers 2 and 3. The detailed training process is explained in Algorithm \ref{alg:Framework}. Coherent with Equation \ref{recon_loss}, $F_{i,j}^k=\{b_i^k,b_j^k\}$ in Algorithm \ref{alg:Framework} is the output reconstructed feature of the $k_{th}$ layer, and $F_{i,j}^{bin}=\{b_i,b_j\}$ is denoted as $F_{i,j}^0=\{b_i^0,b_j^0\}$ for convenience. \\

\vspace{-0.5cm}

After training all the layers, we use the first two encoder layers as the “Compressor”, which is part of the whole model to produce the compressed key feature for the following Siamese verification (as shown in Fig. \ref{whole-fig}(a)).

\begin{algorithm}[!t]
    
    \caption{Hierarchical training of auto-encoders} 
    \label{alg:Framework} 
    \begin{algorithmic}[1]
    \Require Input binarized data pair: denote $F_{i,j}^{bin}=\{b_i,b_j\}$ as $F_{i,j}^0=\{b_i^0,b_j^0\}$; auto-encoder pairs: $AE_{1, 2}=\{AE_1,AE_2\}$, whereas $AE_1^{i}$, $AE_2^{j}$ denotes the $i_{th}$ and $j_{th}$ layer of $AE_1$ and $AE_2$, respectively; the total layer number of the auto-encoder: $L=4$\\ 
    \State \textbf{Initialize:} model parameters $AE_{1, 2}=\{AE_{1, 2}^{(i)},1\leq i\leq L\}$\\
    \For{$k=1; k\leq L/2; k++$}
        \For{$epoch=1; epoch\leq epoch_{max}; epoch++$}
    		\State The encoder forward:
            \For{$q=1; q\leq k; q++$}
    		    \State Forward $F_{i,j}^{q-1}$ on $AE_{1, 2}^q$ batch-wise, obtaining the result $F_{i,j}^{q}$ and retain the gradients.
    	    \EndFor
    		\State Here obtaining $F_{i,j}^{k}$, which is the compressed feature vector.
    		\State The decoder forward:
            \For{$q=L+1-k; q\leq L; q++$}
    		    \State Forward $F_{i,j}^{q-1}$ on $AE_{1, 2}^q$ batch-wise, obtaining the result $F_{i,j}^{q}$ and retain the gradients.
    	    \EndFor
    	    \State Update $AE_{1, 2}^{k}$ and $AE_{1, 2}^{L+1-k}$ with the loss of $\mathcal{L}_{recon}$, while freezing other layer parameters.
    	\EndFor
	\EndFor
    \State Obtain model $AE_{1, 2}$
    \Ensure The stacked encoders $\{AE_{1, 2}^{(i)},1\leq i\leq L/2\}$
    \end{algorithmic}
\end{algorithm}

\subsection{Multi-task learning based Siamese verification model}

The previous feature extraction and compression part of our model focused on distilling longitudinal and morphological information to alleviate noises and imbalance. In this part of the model, we focused on the problem of small sample size. We designed a Siamese verification model as the classifier, based on which the data size can be enlarged from $N$ samples to $[(N-1)\times N]/2$ training pairs. This section corresponds to the “Siamese Verification” part of Fig. \ref{whole-fig}(a). 

Generally speaking, classification is to classify inputs into different category labels, and verification is to determine whether a pair of inputs belong to the same category \citet{Chopra2005Learning}. As for sMRI data with a small number of samples, Siamese verification significantly enlarged the input data amount with cross-combined data pair structure. Rather than extracting traits from each sample, looking for differences between samples is easier for the model to learn from small datasets and complex feature representations. We then use multi-task constraints and brain region weight constraints afterward to validate the use of verification method on the final diagnostic classification.

Therefore, we describe the following three parts of the Siamese verification model in this section. We use the Siamese network structure to enlarge the input sample number, multi-task constraints to optimize the representation space, and the brain region weight constraint mechanism to adapt to the heterogeneity between different samples. 

\subsubsection{Siamese network structure}

Our Siamese verification model is based on the Siamese network \citet{bromley1993signature}, where pairs of data are fed into a pair of networks, so it is necessary to pair the original dataset and reset the label of every data pair. Given $N$ subjects (including both ASD and NC), we can group any two different subjects as one training pair, $F_{i,j}^{cat} = \{n_i, n_j\}$. The labels $y_{ij}$ for paired data are binary, indicating two subjects belonging to the same (defined as $1$) or different (defined as $0$) classes. In this way, we can have $[(N-1)\times N]/2$ training pairs altogether, which is much larger than the number of original subjects.\\

\begin{figure*}
	\centering
	\centerline{\includegraphics[width=.99\textwidth]{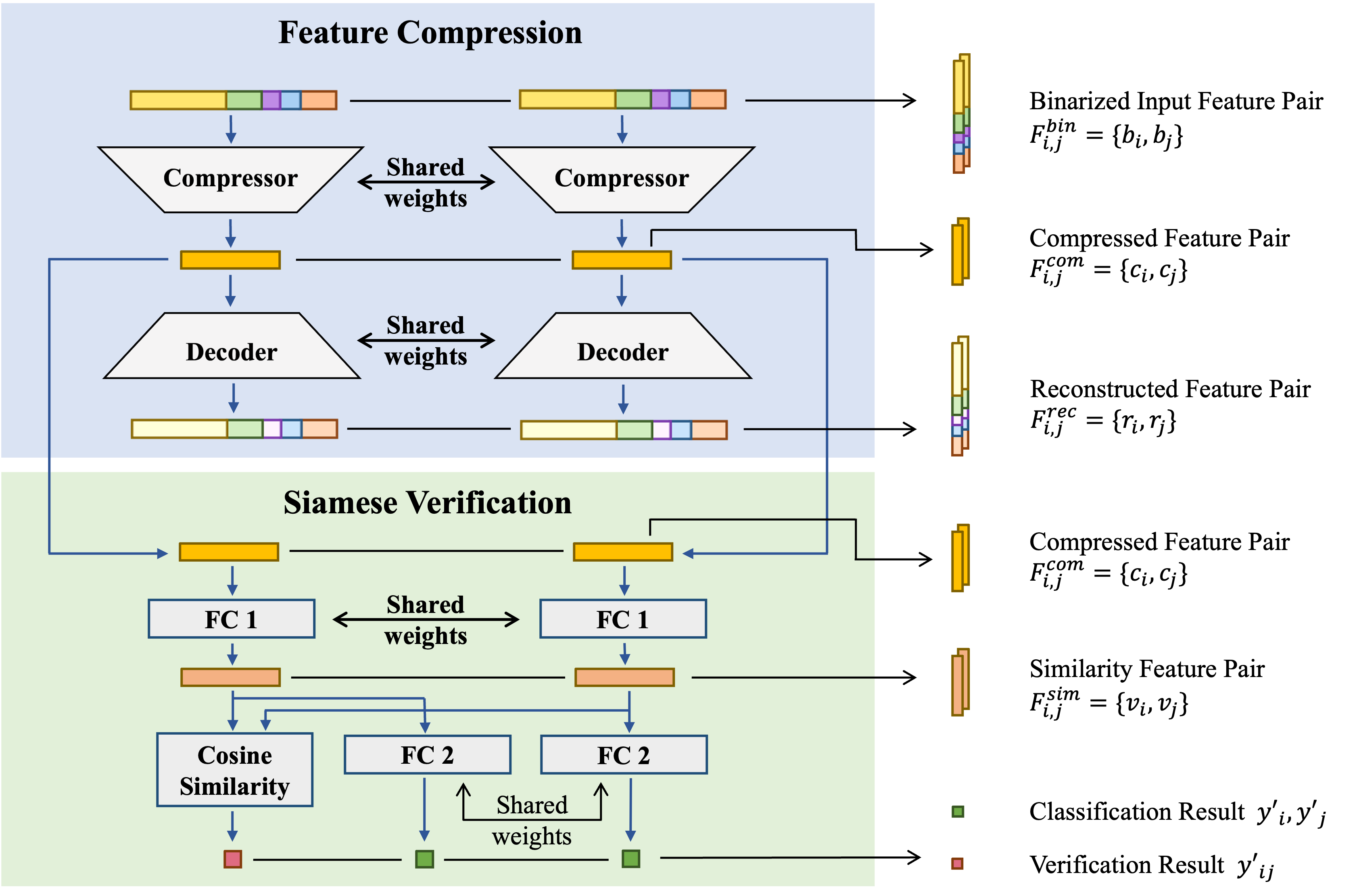}}
	\caption{ 
 The schematic diagram of the multi-task siamese verification model during training. Feature names are listed on the right side of the figure.}
	\label{fig-siamese}
\end{figure*}
\vspace{-0.4cm}

The Siamese verification model takes in the output of the dual-channel autoencoder. We use the compressed feature $F_{i,j}^{com}$ produced by the Compressor as the Siamese verification input. A fully-connected layer FC1 is added to further produce the similarity feature $F_{i,j}^{sim}$, as shown in Fig. \ref{fig-siamese}. In this way, effective features can be obtained through the unsupervised trained encoder module of the dual channel autoencoder, while the additional FC can help to capture the feature information, which is more conducive to the classification. After the dual-channel network layers, a similarity measurement function $\mathcal{S}$ is defined as the cosine similarity, 

\begin{equation}
\begin{split}
\mathcal{S}(v_i,v_j)=\dfrac{v_i\cdot v_j}{{\left \| v_i \right \|}{\left \| v_j \right \|}}
\end{split}
\label{simequ}
\end{equation}

where $F_{i,j}^{sim} = \{v_i, v_j\}$ are FC outputs of the compressed feature $F_{i,j}^{com}$. The distance between two features would be small if they belong to the same classes and large otherwise.

\subsubsection{Multi-task constraints}

According to the previous description, we relabel paired samples into pairs of “same” or “different” types of samples, denoted by “1” and “0”, respectively. However, pairs of ASD and pairs of normal samples would both be labeled as the “same” type. The use of Siamese verification has changed the original problem space from a classification to a verification one, which will have a negative impact on the final classification. We therefore use an additional classification loss to optimize the learning process by further expanding inter-class distances.

 In order to alleviate the influence on the similarity feature $F_{i,j}^{sim}={v_i,v_j}$ in the representation space, we propose a multi-task learning constraint framework. Through multi-task learning, we seek to narrow the spatial distance of sample features with the same diagnostic labels in the representation space, meanwhile further expanding the distance between features with different diagnostic labels by adding classification loss. Specifically, the original labels of a pair of input samples are introduced as the training target of two other single branches, as shown in Fig. \ref{fig-siamese}. Through this multi-task constraint framework, the “Verification Result” on the left of Fig. \ref{fig-siamese} is used for verification, while the two branches on the middle and the right side of Fig. \ref{fig-siamese} would do classification. 

Suppose the input paired data is $F_{i,j}^{sim}=\{v_i,v_j\}$, then $y_i$ and $y_j$ denote the classification labels of sample $i$ and $j$.  The loss function of multi task constraints $\mathcal{L}_{mul}$ is as follows:

\begin{equation}
\mathcal{L}_{mul} = \mathcal{L}_{ver}+\mathcal{L}_{cls}
\end{equation}

For the classification loss $\mathcal{L}_{cls}$, we use $y'_i$ and $y'_j$ to denote the classification results of sample $i$ and $j$ with the Siamese verification model, as shown in Fig. \ref{fig-siamese}. The classification loss $\mathcal{L}_{cls}$ is defined as:

\begin{equation}
\begin{split}
\mathcal{L}_{cls} =&  \text{CEL}(y_i,y'_i)+\text{CEL}(y_j,y'_j)\\
=&-(y_ilog{y'_i}+(1- y_i)log(1-y'_i))\\
&-(y_jlog{y'_j}+(1- y_j)log(1-y'_j))
\end{split}
\end{equation}

where CEL denotes cross-entropy loss function.

For the verification loss $\mathcal{L}_{ver}$, we define $y_{ij}$ and $y'_{ij}=\mathcal{S}(v_i,v_j)$ to denote the label and the similarity result of the input paired sample ${i,j}$. The verification loss $\mathcal{L}_{ver}$ is defined as:

\begin{equation}
\mathcal{L}_{ver}=\text{FL}(\hat{y}_{ij}) =-\alpha(1-\hat{y}_{ij})^\gamma log(\hat{y}_{ij})
\end{equation}

where FL denotes the focal loss function \citet{lin2017focal}, with an intermediate variable:

\begin{equation}
\hat{y}_{ij}=\left\{
\begin{aligned}
y'_{ij} & , & if y_{ij}=1 \\
1- y'_{ij} & , & otherwise
\end{aligned}
\right.
\end{equation}

\subsection{Test Phase: brain region weight constraint mechanism}

\begin{figure}[htb]
	\centering
	\centerline{\includegraphics[width=8.5cm]{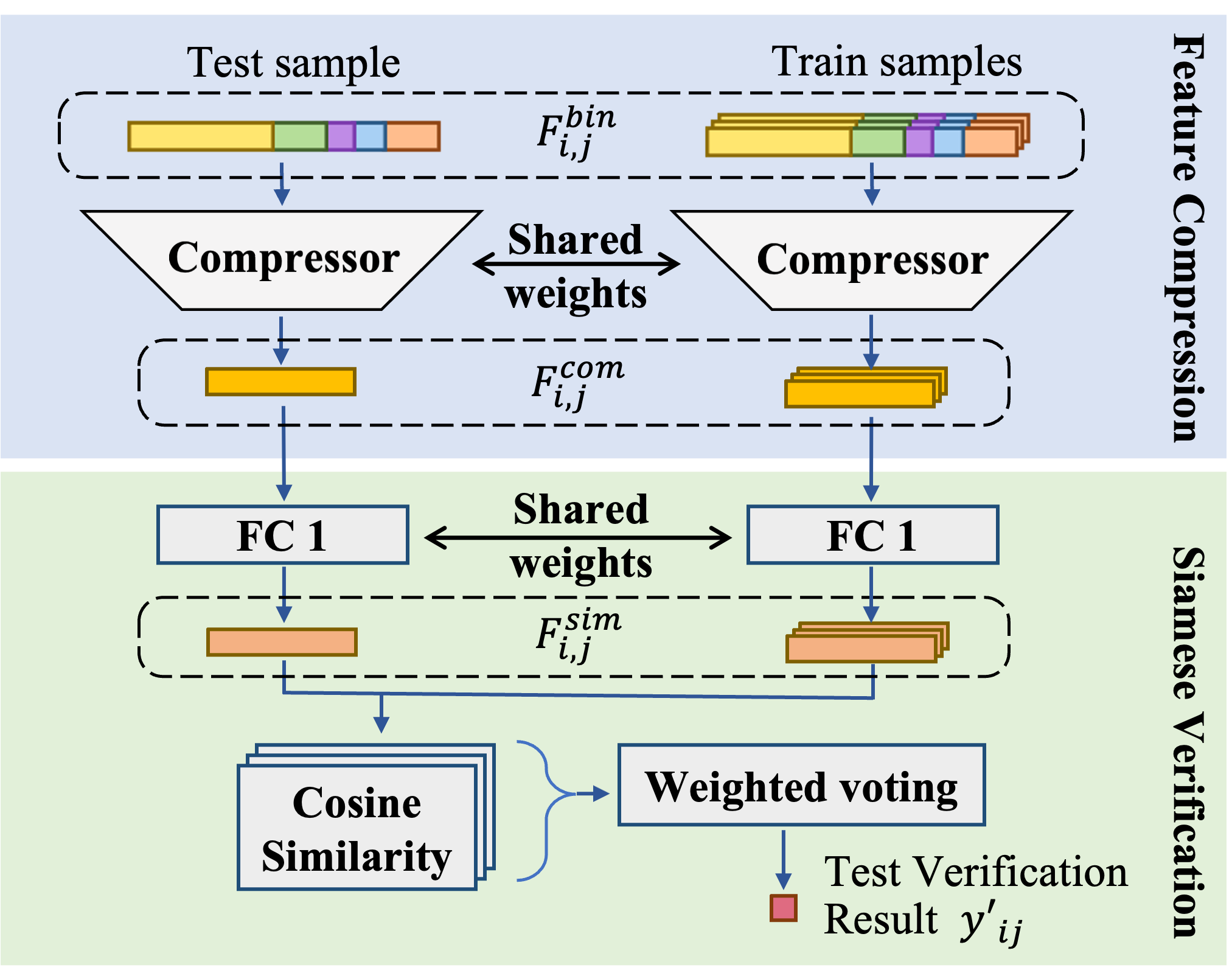}}
	\caption{ 
 Test phase of the Siamese verification model. The test similarity result is obtained by verification with all training samples and the voting afterwards. Classification modules are only used in training phase to calculate the integrated loss and are not utilized in test phase.}
	\label{fig-siamese-test}
\end{figure}

In the test phase of the dual-channel verification model, the final similarity result is obtained through voting, in which every training sample would be used to calculate a similarity score with the currently tested sample. Nonetheless, early brain development is subject-specific, so the referential importance of each training sample in different test trials should be correspondingly different. In order to make the similarity vote more reasonable in the test stage, voting weight values are introduced by using the similarity information of the binarized feature vector. This part is consecutive to the cosine similarity module, denoted by “Weighted voting” in Fig. \ref{fig-siamese-test}.\\

\vspace{-0.4cm}

 In order to categorize a test sample $i$, all samples in the training dataset $D_t$ are paired with $i$ to do weighted voting. The training dataset consists of autistic infant samples $D_a$ and normal infant samples $D_n$. Therefore, after pairing the test sample with training samples, the similarity of the test sample $i$ with autistic samples $D_a$ and normal samples $D_n$ can be obtained respectively. For the sample $i$ to be tested, the final classification result is determined by comparing the similarity score between the autistic infant similarity $\mathcal{S}_a$ and the normal infant similarity $\mathcal{S}_n$, the larger one would be the final category of the test sample. The specific calculation formula is as follows:

\begin{equation}
    \mathcal{S}_a = \dfrac{\sum_{j=1}^{|D_a|}w_j\mathcal{S}(v_i,v_j)}{|D_a|} ,  j \in D_a
\end{equation}

\begin{equation}
    \mathcal{S}_n = \dfrac{\sum_{j=1}^{|D_n|}w_j\mathcal{S}(v_i,v_j)}{|D_n|} ,  j \in D_n
\end{equation}

 where $|D_a|$ and $|D_n|$ denote the sample numbers of $D_a$ and $D_n$, respectively. $i$ and $j$ denote the test sample and the training sample of the current sample pair. $w_j$ represents the similarity weight constraint of the input binarized feature pair vectors,
 
\begin{equation}
    w_j = \mathcal{S}(b_i,b_j) ,  j \in D_t
\end{equation}

where $\mathcal{S}$ denotes the similarity measurement function defined in Equation \ref{simequ}. As shown in Fig. \ref{fig-siamese}, $F_{i,j}^{bin}=\{b_i,b_j\}$ is the binarized feature pair, $F_{i,j}^{sim}=\{v_i,v_j\}$ denotes the output Siamese feature of the fully connected network. 
 
\subsection{Region importance extraction from model weights}

\begin{figure}[htb]
	\centering
	\centerline{\includegraphics[width=8.5cm]{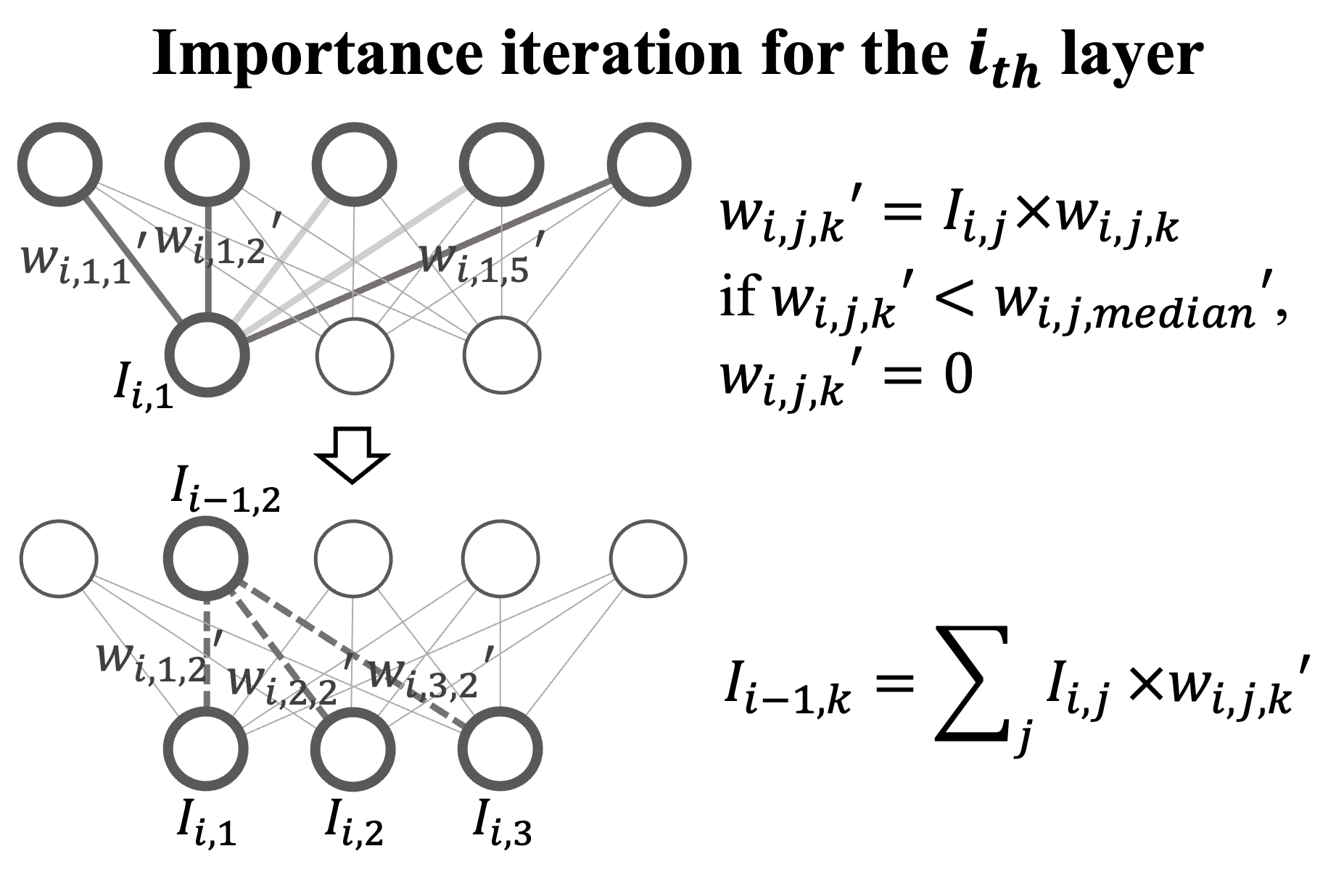}}
	\caption{ 
 The importance iteration for the $i$-th layer. Step 1: calculate $w_{i,j,k}'$ with the importance factor $I_{i,j}$ and set the value of those below the median to 0. Step 2: calculate the previous layer's importance factor $I_{i-1,k}$ with the weighted sum of $I_{i,j}$ and $w_{i,j,k}'$.}
	\label{fig-imp-iter}
\end{figure}

We use an iterative method to extract the importance of each input feature dimension from the final auto-encoder module, and obtain the importance of each brain region in our model's diagnosis, as shown in Fig. \ref{fig-imp-iter}. Since our network input is binarized, all nodes share the same scale and we can use their absolute parameter value to represent their importance, which is similar to \citet{hazlett2017early}. We use $w_{i,j,k}$ to denote the absolute value of the weight corresponding to the $j_{th}$ output neuron and the $k_{th}$ input neuron in the $i_{th}$ layer of our auto-encoder. The $j_{th}$ output neuron of the $i_{th}$ layer (which equals to the $j_{th}$ input neuron of the ${i+1}_{th}$ layer) is given an importance factor $I_{i,j}$, to represent the importance of this position. We set the importance factor of the last layer's output equally to 1. Then for each layer, we follow the equations below to calculate the importance factor of its previous layer in two steps.\\

\vspace{-0.4cm}

Step 1, we select important connections by multiplying the importance factor, and discarding weights below the median. Namely,

\begin{equation}
{w_{i,j,k}}' = I_{i,j}\times w_{i,j,k} ; \text{if }{w_{i,j,k}}'<{w_{i,j,median}}' , {w_{i,j,k}}'=0
\label{eq_impor_1}
\end{equation}

where ${w_{i,j,k}}'$ is the weighted network weight, ${w_{i,j,median}}'$ is the median of all ${w_{i,j,k}}'$ connected to $I_{i,j}$.

Step 2, we then integrate the previous layer's importance factor with

\begin{equation}
I_{i-1,k} = \sum_{j} I_{i,j} \times {w_{i,j,k}}'
\label{eq_impor_2}
\end{equation}

We do this calculation from the last layer to the first layer, to obtain the importance factor $I$ of each of the input feature dimensions. We then combine those importance factors of both cortical features and cortical PS features to obtain the cortical region importance.

\section{Experiments and results}
\subsection{Experimental setting}
The experimental data in this article is the infant structural magnetic resonance imaging data from the NDAR data-base, including data from 30 autistic infants and 127 normal controls. We evaluate the method with 10-fold cross-validation strategy. Therefore, during training, there are 141 training samples and 16 test samples in each fold. 

The dual-channel autoencoder is composed of a two-layer Compressor and a two-layer Decoder. The encoder compresses the input feature from 1,265 dimensions to 128 dimensions in its first layer and compresses it further into 16 dimensions in its second layer, while the decoder decodes the vector from 16 dimensions to 128 dimensions and then reconstructs the 1,265-dimensional feature. The Siamese verification model network is based on the 16-dimensional vector output by the Compressor, adding a layer of compression network to compress the data to 4 dimensions and then calculate the cosine similarity of the paired input data. 

In order to objectively and accurately validate the model, this article evaluates the model from four aspects: accuracy, sensitivity, specificity, and F1-score. Accuracy refers to the proportion of the number of samples that are correctly predicted in the entire test set; sensitivity (\ie\ recall) refers to the proportion of positive (autism) samples correctly detected; specificity refers to the proportion of negative (normal) samples correctly classified; and F1-score is a comprehensive indicator of the model performance, which is the harmonic mean of precision (the number of true positive predictions divided by the number all positive predictions) and sensitivity (recall). We report the highest results of finetuned models. In the process of clinical diagnosis, it is particularly important to be able to accurately identify autism samples, so the F1-score and sensitivity are of more significance for evaluating the model’s diagnosis ability than accuracy alone.

\subsection{Ablation experiments}

In this section, we conduct ablation experiments to evaluate the effectiveness of our model. In order to prevent over-fitting of unsupervised training, the concatenated feature is binarized before entering the dual-channel autoencoder network. The ablation experiment results shown in Table \ref{tab:binarization} compare the model's performance when using the binarized features versus the original normalized features. It can prove that the performance of the algorithm after binarization is better, where the autistic recalling ability of the model is decreased without binarization. This result may indicate that the key structural and developmental information for autism diagnosis is relatively invariant during binarization.

\begin{table}[h]
    \centering
    \caption{ 
    The model effect with \& without binarization operation, evaluated with 10-fold cross-validation.}
\begin{tabular}{l|ccccc}
    \toprule
Model  &  F1-score & ACC & SEN & SPE \\
\hline
w/o binarization& 0.516 & 0.808 & 0.533 & 0.874   \\
Binarization (Ours)  & {\bf 0.585} & {\bf 0.829} & {\bf 0.633} & {\bf 0.874}  \\
    \botrule
\end{tabular}\vspace{0cm}
    \label{tab:binarization}
\end{table}

Moreover, explorations have been done on the weighted voting module in the testing phase, with comparison results shown in Table \ref{tab:weight}. The first attempt is to directly calculate the average similarity result without using weight constraints to consider the heterogeneity between samples, denoted by “W/o weight”. The second one is to consider the heterogeneity between samples and use sample compressed feature vector $F_{i,j}^{com} = \{c_i, c_j\}$ produced by the Compressor as the weight constraint metrics, denoted by “Comp\_weight”. The third one is to use the binarized input feature vector $F_{i,j}^{bin} = \{b_i, b_j\}$ as the weight constraint metrics, denoted by “Bi\_weight”. The comparison of the results shows that the “Bi\_weight” weight constraint method performs best, due to its advantages in representing the complete original features for sample similarity constraints.

\begin{table}[h]
    \centering
    \caption{ 
    The model effect with different brain region weight constraint mechanisms, evaluated with 10-fold cross-validation.}
\begin{tabular}{l|ccccc}
    \toprule
Model &  F1-score & ACC & SEN & SPE  \\
\hline
w/o weight  & 0.540 &  0.816&  0.566 & 0.874  \\
Comp\_weight  & 0.545 & 0.810 & 0.600 & 0.858 \\
Bi\_weight (Ours)  & {\bf 0.585} & {\bf 0.829} & {\bf 0.633} & {\bf 0.874} \\
    \botrule
\end{tabular}\vspace{0cm}
    \label{tab:weight}
\end{table}

Ablation experiments of each module are also carried out to verify the corresponding module effectiveness, with results shown in Table. \ref{tab:ablation}. “Ours” denotes our algorithm, which is deploying all modules. “W/o weight \& AE” denotes the algorithm obtained on the basis of the Siamese verification network with PS feature as \citet{zhang2020siamese}, where neither dual-channel autoencoder modules nor brain region weight constraint mechanism modules is used. In a similar way, “W/o weight” denotes the algorithm after removing the brain region weight constraint module in the test phase. “W/o PS” denotes an algorithm that does not add path signature features. “W/o AE” denotes an algorithm that does not add the dual-channel autoencoder module. 

By comparing the “W/o weight” and the original model, it could be known that the brain region weight constraint mechanism, which determines the weight of each voting sample pair during the test phase, can effectively further improve model sensitivity. The “W/o AE” experiment shows that it is meaningful to use the feature information of the brain to perform feature learning and compression in an unsupervised manner. The comparisons between the results of “W/o PS” and the original model also illustrate that the path signature module is crucial for feature extension and extraction. Besides, comparing the results of “W/o weight \& AE” and the original model, we evaluate the comprehensive performance of the model.

It is worth noting that although the accuracy increase of our full algorithm is not that large in Table \ref{tab:ablation} compared to other ablation models, there is a significant difference in  sensitivity and F1-score, which better represents the comprehensive performance of our model. This is because a model with high accuracy and low sensitivity is essentially overfitting to negative samples under an imbalanced dataset, thus learning much less from data than models with higher F1-scores.

\begin{table}[h]
    \centering
    \caption{ 
    Ablation experiments of each module of our model  with 10-fold cross-validation.}
\begin{tabular}{l|cccc}
    \toprule
Model  &  F1-score & ACC & SEN & SPE \\
\hline
w/o weight \& AE  & 0.348 & 0.810& 0.266 & 0.937 \\
w/o PS  & 0.526 & 0.829 & 0.500 & 0.926 \\
w/o AE  & 0.375 & 0.810 & 0.300 & 0.929 \\
w/o weight  & 0.540 & 0.816 &  0.566 & 0.874 \\
Ours  & {\bf 0.585} & {\bf 0.829} & {\bf 0.633} & {\bf 0.874} \\
    \botrule
\end{tabular}\vspace{0cm}
    \label{tab:ablation}
\end{table}

\subsection{Comparison with other methods}

\begin{figure}[htb]
	\centering
	\centerline{\includegraphics[width=7.6cm]{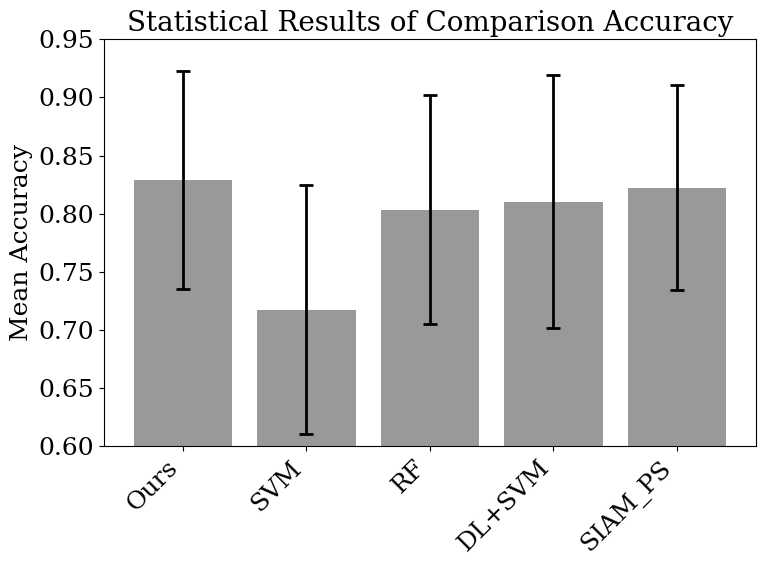}}
	\vspace{-0.2cm}
	\caption{ 
 Visualizing the statistical results of accuracies of different methods. Error bars indicate standard deviations.} 
	\label{fig-sta-compare}
\end{figure}

Finally, we compare our own algorithm with other related works, as shown in Table \ref{tab:whole}. SVM denotes the Support Vector Machine (SVM) classification method. RF denotes the random forest method. SIAM\_PS is the work of \citet{zhang2020siamese}, and DL+SVM refers to the work of \citet{hazlett2017early}. “Our’s” denotes our algorithm. All the comparison experiments are conducted under the same conditions on the same dataset, as introduced in the Materials section. Since the imbalance in the number of data categories is a major characteristic of the dataset, the compared algorithms generally have poor recall rates, which means the corresponding sensitivity value is very low in Table \ref{tab:whole}. The DL+SVM method have even 0 recall with the categorically imbalanced dataset. By comparing our results with other algorithms, we can find that our model is specifically effective for the recall of autism samples, which increases the model sensitivity while maintaining high accuracy, represented by a high F1-score. We also include Fig. \ref{fig-sta-compare} to better present the statistical property of our comparison experiment results visually. It is shown that, although our results have some fluctuations due to small sample size, standard deviations are in reasonable ranges.

\begin{table}[h]
    \centering
    \caption{ 
    Comparison with other autism diagnosis models under few-shot and class-imbalanced conditions, on the same dataset with 10-fold cross-validation.}
\begin{tabular}{l|cccc}
    \toprule
Model  &  F1-score & ACC & SEN & SPE \\
\hline
SVM & 0.241 & 0.718 & 0.233 & 0.834 \\
RF  & 0.057 & 0.791 & 0.033 & 0.965 \\
DL+SVM & 0.000 & 0.810 & 0.000 & 1.000  \\
SIAM\_PS  & 0.391 & 0.823 & 0.300 & 0.945  \\
Ours & {\bf 0.585} & {\bf 0.829} & {\bf 0.633} & {\bf 0.874}  \\
    \botrule
\end{tabular}\vspace{0cm}
    \label{tab:whole}
\end{table}

Additionally, we summarize some recent  works using MRI data for autism classification in Table \ref{tab:review}, to provide a broader view on the progress of the field. Of note, those various works used very different input data types and datasets with different age ranges, so this is not a quantitively fair comparison, but only an additional general review. Works listed in the table utilized different imaging modalities and computational methods for autism diagnosis. Compared with other works, our method performs well under practical scenarios of scarce, imbalanced, infantile, and sMRI-only data.

\begin{table*}[t]
    \centering
    \caption{ 
    A brief summary of recent works on autism classification with MRI data. The short forms used in the table are: ASD, Autism Spectrum Disorder; DL, Deep Learning; DTI, Diffusion Tensor Imaging; FCM, Functional Connectivity Matrices; MBN, Morphological Brain Networks; NC, Normal Controls; RF, Random Forests; SVM, Support Vector Machines.}
\begin{tabular}{l|cccccccc}
    \toprule
Reference  & \makecell{Sample Size \\ (ASD/NC)} & Data Type & Model \& Method & \makecell{Age \\ (years)} &  F1-score & ACC & SEN & SPE\\
\hline
\citet{zhao2018diagnosis} & 100 (54/46) & fMRI & multi-order FCMs & under 15 & 0.83 & 0.81 & 0.82 & 0.80 \\
\citet{soussia2018unsupervised} & 341  (155/186) & sMRI & multi-order MBNs & $\sim$ 17 & - & 0.6170 & - & - \\
\citet{zhang2018whole} & 149 (70/79) & DTI & SVM & 6-18 & - & 0.7833 & 0.8481 & 0.7286 \\
\citet{sen2018general} & 1111 (538/573) & fMRI+sMRI & DL+SVM & 7-64 & - & 0.6431 & 0.6832 & 0.6000 \\
\citet{li2018early} & 264 (55/209) & sMRI & DL & 2 & - & 0.7924 & - & - \\
\citet{kong2019classification} & 183 (78/104) & sMRI & DL & $\sim$ 15 & - & 0.9039 & 0.8437 & 0.9588 \\
\citet{rakic2020improving} & 817 (368/449) & fMRI+sMRI & DL & 7-64 & - & 0.8445 & 0.80 & 0.88 \\
\citet{kim2022classification} & 106 (58/48) & sMRI+DTI & RF & 3-6 & - & 0.888 & 0.930 & 0.838 \\
\citet{gao2022deep} & 1994 (N/A) & sMRI & DL & 5-64 & - & 0.6785 & 0.6166 & 0.7336 \\
Ours & 157 (30/127) & sMRI & DL & 0-1 & 0.585 & 0.829 & 0.633 & 0.874 \\
    \botrule
\end{tabular}\vspace{0cm}
    \label{tab:review}
\end{table*}

\subsection{Cortical region importance analysis}

\begin{figure}[htb]
	\centering
	\centerline{\includegraphics[width=8.2cm]{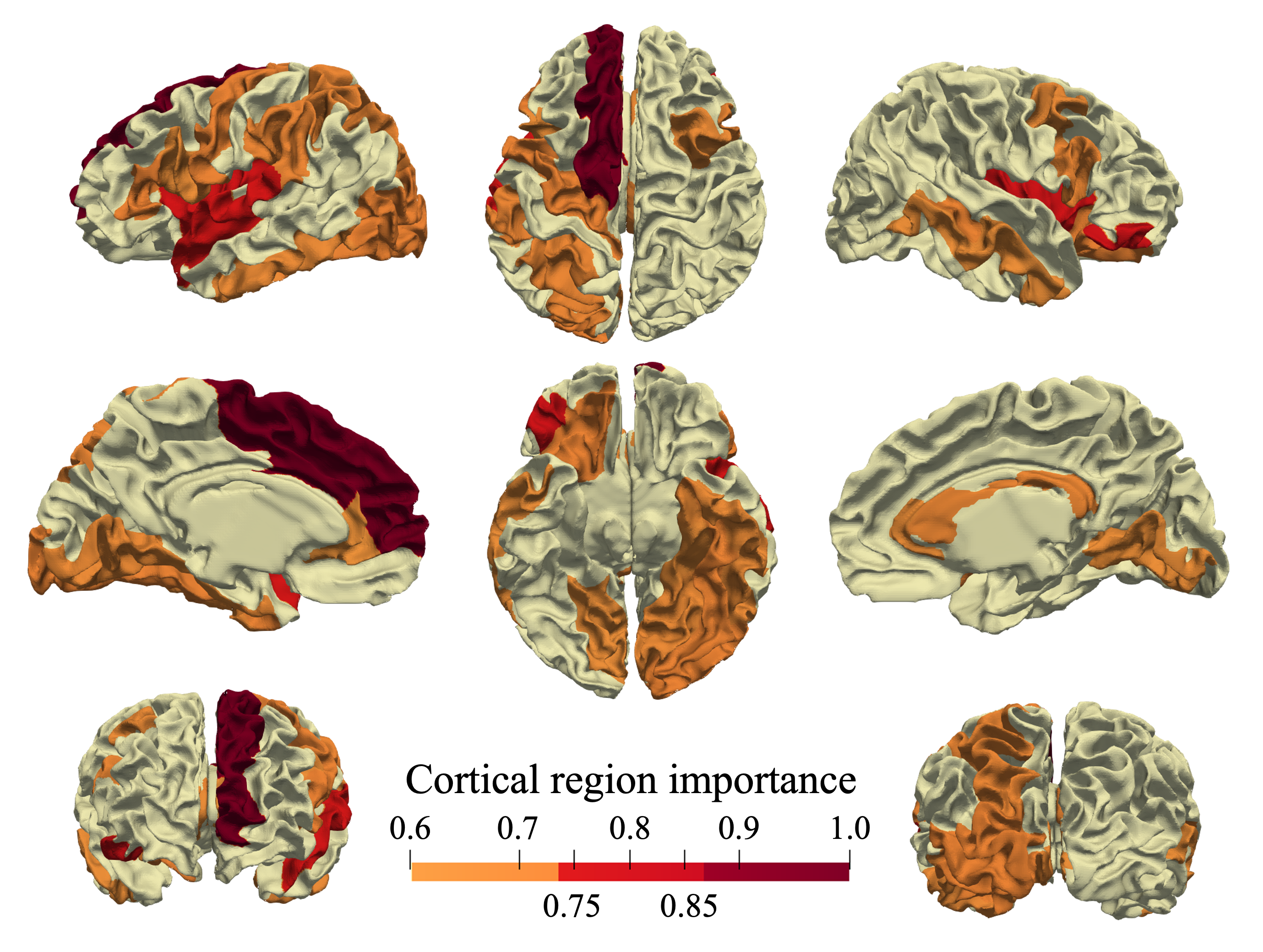}}
	\caption{ 
 Visualization of cortical regions’ influence on the final classification results. The top 20 regions (out of 70 regions) are extracted from the final deep learning model and visualized. Colors of the cortical regions indicate their normalized regional importance, as indicated above. The top 6 cortical regions include: the left superior frontal gyrus, the left caudal anterior-cingulate cortex, the right bank of the superior temporal sulcus, the left superior temporal gyrus, the left fusiform gyrus, and the right lateral orbital frontal cortex. }
	\label{fig-cortical-region}
\end{figure}

Our input features of the model are based on the infant dedicated cortical surface atlas proposed by \citet{wu2019construction}, from which we used features of 70 cortical ROIs. The importance of each brain region is calculated with the method described in the Region importance extraction part of the Method section, and the importance is normalized into the range of 0.2 to 1 for better visualization. Fig. \ref{fig-cortical-region} presents the visualization of the most influential cortical regions during the classification. 

As shown in the figure, the left superior frontal gyrus, the left caudal anterior-cingulate cortex, the right bank of the superior temporal sulcus, the left superior temporal gyrus, the left fusiform gyrus, and the right lateral orbital frontal cortex are regions considered as brain regions containing the most important predictive information by our model. Our reported machine-learning-focused cortical regions in infants may reveal underlying information regarding ASD early development, and be indicative for future research.

We also analyzed and included the importance of the gender features following the same extraction and normalization method in the Region importance extraction part of the Method section, and compared them with brain regional features. The normalized gender feature importance is 0.170, which is less than all the brain region importance weights ranging from 0.2 to 1, indicating that our model is paying less attention to gender information, compared to each of the brain regional MRI features. 

However, since our gender feature input is 2-dimensional and the MRI features of each brain region are expanded to 18-dimensional, following the feature extraction process described in the Methods section, the importance comparison between gender and brain regional features may be biased even after normalization. So we additionally conducted ablation experiments with and without gender information as an input feature. For the “w/o gender info” experiments, we remove the gender features in the input and keep all other features. The ablation experiment results are shown in Table 6 below. The results showed that removing gender information negatively affects the model performance. This ablation experiment indicates that gender information is a helpful supplement to image data in autism diagnosis.

\begin{table}[h]
    \centering
    \caption{ 
    The model performance with and without gender information, evaluated with 10-fold validation.}
\begin{tabular}{l|ccccc}
    \toprule
Model  &  F1-score & ACC & SEN & SPE \\
\hline
w/o gender info& 0.484 & 0.797 & 0.500 & 0.866   \\
With gender info (Ours)  & {\bf 0.585} & {\bf 0.829} & {\bf 0.633} & {\bf 0.874}  \\
    \botrule
\end{tabular}\vspace{0cm}
    \label{tab:gender}
\end{table}

\section{Discussion and conclusions}
In this paper, we proposed a class-imbalanced deep learning scheme based on Siamese verification and unsupervised feature compression for ASD diagnosis from sMRI data in infants. Our work mainly focused on addressing the realistic problems of small sample size, categorical imbalance, cortical developmental heterogeneity between different samples, and complexity of longitudinal data. Compared with other works, the use of Siamese verification in our work substituted classification with pairing, thus augmenting the dataset size significantly. We also adapted Path Signature and unsupervised stacked autoencoders to further compress and extract sMRI features, hence improving the classification performance on imbalanced datasets, especially on sensitivity. Besides, we proposed the brain region weight constraint mechanism, to enhance the classification voting process with regional similarity attention weight. 

It is noted that many studies of autism diagnoses from MRI data are based on functional MRI, like what was proposed and compared by \citet{sherkatghanad2020automated}, which may be partly due to the convenience brought by the public ABIDE dataset \citep{di2014autism}. However, regarding the specific problem of infant autism prediction, there have been more studies on the structural developmental abnormalities of autistic infants than functional abnormalities \citep{dawson2022prediction}. Therefore, our work uses structural MRI data to predict autistic infants, and we specifically utilize Path Signature to decode structural developmental abnormalities. Nevertheless, there has also been infant autism diagnosis from functional MRI \citep{emerson2017functional}, and it would be interesting to explore the combination of structural and functional MRI data on this problem by leveraging their complementary information.

Regarding critical brain regions for autism diagnosis, our method indicated that the left superior frontal gyrus, the left caudal anterior-cingulate cortex, the right bank of the superior temporal sulcus, the left superior temporal gyrus, the left fusiform gyrus, and the right lateral orbital frontal cortex are the most important regions, in line with some related works. For example, \citet{hazlett2017early} used a machine learning model to obtain the most important brain regions for autism prediction and found that the left superior frontal gyrus, the left fusiform gyrus, and the right orbital frontal cortex were within the top 8 contributive brain regions out of 78 regions in total. Moreover, \citet{wang2022developmental} studied structural covariance networks of cortical thickness and surface area in autistic infants within the first two years to detect morphological and developmental abnormalities via statistical analysis. They also found the left superior frontal gyrus, the left superior temporal gyrus, and the right medial orbital frontal gyrus show abnormalities. And they reported a developmental abnormality in the left occipital cortex as well, which is another focused region in our model. These results indicate that our model has extracted meaningful key features for autism diagnosis.

Our model focused the most on the left superior frontal gyrus and the left caudal anterior-cingulate cortex for autism prediction, largely overlapping with the medial prefrontal cortex (mPFC). While there are few reports on structural abnormalities in mPFC, a lot of work indicated mPFC functional abnormalities in autism patients. \citet{shalom2009medial} summarized the functional role of mPFC in autism, showing that the integrative level abnormalities of emotion, memory, sensation-perception, and motor skills in mPFC are associated with autism symptoms. \citet{gilbert2008atypical} also demonstrated mPFC functional abnormalities in executive function tasks. These functional highlights of mPFC in autism are consistent with our results on structural MRI data.

The right superior temporal sulcus (STS) and the left superior temporal gyrus (STG) are another two regions that our model focused on. They have also been identified by other studies as regions closely related to autism. For STS, \citet{redcay2008superior} discussed its functions of social and speech perception, and argued that social and language related to autism symptoms originated from STS functional impairments. More directly, \citet{boddaert2004superior} observed anatomical abnormalities in STS from autistic children, which aligns with our findings from structural MRI. For STG, several works particularly addressed the failure of the left STG lateralization in the early development of autism \citep{bigler2007superior,eyler2012failure}, supporting our results’ focus on the early development abnormalities in the left STG. Taken together, the tight connections of mPFC, STS, and STG abnormalities with autism have been demonstrated by previous literature. The ability of our model to automatically focus on those key regions from scarce, imbalanced data for autism diagnosis, has proven the effectiveness of our model and training procedure design.

As for future improvements to our work, we have the following plans. (1) Our work adapted sMRI data for classification. As mentioned previously, combining other forms of data, like diffusion MRI and functional MRI data, may provide more relevant implications. (2) Since human brains can be represented by graph connectivity, cortical graph modeling and classification with graph convolutional networks (GCNs) may have advantages in feature extraction. (3) Regarding the feature transformation using Path Signature, learnable Path Signature features may complement our feature extraction.

\section*{Competing interests}

The authors declare that they have no known competing financial
interests or personal relationships that could have appeared to influence the work reported in this paper.

\section*{Data Availability Statement}

Data used in the preparation of this manuscript were obtained from the NIH-supported National Database for Autism Research (NDAR). NDAR is a collaborative informatics system created by the National Institutes of Health to provide a national resource to support and accelerate research in autism. This manuscript reflects the views of the authors and may not reflect the opinions or views of the NIH or of the submitters submitting original data to NDAR.

\section{Author contributions statement}

Zhuowen Yin (Conceptualization, Formal Analysis, Investigation, Methodology, Resources, Software, Validation, Visualization, Writing – original draft, Writing – review \& editing), Xinyao Ding (Conceptualization, Methodology, Software, Writing – original draft), Xin Zhang (Conceptualization, Methodology, Project administration, Funding acquisition, Resources, Supervision, Writing – review \& editing), Zhengwang Wu (Data curation, Writing – review \& editing), Li Wang (Data curation, Resources), Xiangmin Xu (Conceptualization, Project administration, Funding acquisition, Resources, Supervision), and Gang Li (Conceptualization, Data curation, Project administration, Resources, Supervision, Writing – review \& editing).

\section{Acknowledgments}
This work was supported by the Guangdong Provincial Key Laboratory of Human Digital Twin (2022B1212010004 to X. Z. and X. X.), and the Fundamental Research Funds for the Central Universities (No. 2022ZYGXZR104 to X. Z.).



\bibliographystyle{abbrvnat}
\bibliography{refs}

\end{document}